\documentclass[conference]{IEEEtran}
\IEEEoverridecommandlockouts

\usepackage{amsmath,amssymb,amsfonts}
\usepackage{algorithmic}
\usepackage{graphicx}
\usepackage{textcomp}
\usepackage{xcolor}
\usepackage{multirow}
\usepackage{booktabs}
\usepackage{subcaption}
\usepackage{float}
\usepackage{hyperref}
\usepackage{url}

\makeatletter
\renewenvironment{abstract}{%
    \par\noindent{Abstract: }\ignorespaces
}{\par}
\makeatother
\def\BibTeX{{\rm B\kern-.05em{\sc i\kern-.025em b}\kern-.08em
    T\kern-.1667em\lower.7ex\hbox{E}\kern-.125emX}}

\begin{document}

\title{RoboLayout: Differentiable 3D Scene Generation for Embodied Agents}

\author{
\IEEEauthorblockN{Ali Shamsaddinlou\thanks{Code available at \url{https://github.com/alishams21/robolayout}}}
}

\maketitle
\maketitle

\begin{abstract}
Recent advances in vision–language models (VLMs) have shown strong potential for spatial reasoning and 3D scene layout generation from open-ended language instructions. However, generating layouts that are not only semantically coherent but also feasible for interaction by embodied agents remains challenging, particularly in physically constrained indoor environments. In this paper, RoboLayout is introduced as an extension of LayoutVLM that augments the original framework with agent-aware reasoning and improved optimization stability. RoboLayout integrates explicit reachability constraints into a differentiable layout optimization process, enabling the generation of layouts that are navigable and actionable by embodied agents. Importantly, the agent abstraction is not limited to a specific robot platform and can represent diverse entities with distinct physical capabilities, such as service robots, warehouse robots, humans of different age groups, or animals, allowing environment design to be tailored to the intended agent. In addition, a local refinement stage is proposed that selectively re-optimizes problematic object placements while keeping the remainder of the scene fixed, improving convergence efficiency without increasing global optimization iterations. Overall, RoboLayout preserves the strong semantic alignment and physical plausibility of LayoutVLM while enhancing applicability to agent-centric indoor scene generation, as demonstrated by experimental results across diverse scene configurations.
\end{abstract}

\section{Introduction}

3D scene generation aims to synthesize spatially structured, semantically coherent, and physically plausible environments for applications such as robotics \hyperref[sec:references]{[2]}, AR/VR \hyperref[sec:references]{[3]}, world modeling \hyperref[sec:references]{[4]}, and architectural visualization \hyperref[sec:references]{[5]}. A fundamental challenge in this domain lies in aligning high-level semantic intent expressed through natural language with low-level geometric, spatial, and agent-specific constraints that govern object placement and scene structure.

Recent progress in vision–language models (VLMs) and differentiable optimization has enabled new approaches to this problem. LayoutVLM \hyperref[sec:references]{[1]} represents a significant advance by formulating 3D layout generation as a differentiable optimization process guided by vision–language supervision. By integrating semantic reasoning with gradient-based refinement, LayoutVLM achieves improved alignment between textual descriptions and spatial layouts.

In this paper, RoboLayout is proposed as an extension of LayoutVLM that incorporates agent-aware reachability and a post-optimization cleaning process. First, explicit reachability constraints are integrated into the differentiable optimization loop, enabling the generation of layouts that are feasible and actionable for a broad class of embodied agents. Second, a local refinement stage selectively re-optimizes only problematic object placements while keeping the remainder of the layout fixed, improving convergence efficiency without increasing global iteration counts. This combination allows RoboLayout to disentangle global structural objectives from local relational refinement while maintaining semantic consistency.

By building directly on the LayoutVLM framework, RoboLayout preserves end-to-end differentiability and compatibility with vision–language supervision, and importantly, the agent abstraction enables layouts to be explicitly tailored to the intended actor’s physical capabilities, bringing environment design closer to practical deployment for diverse real-world scenarios.

\section{Contributions}

The key contributions of this work are summarized as follows:

\begin{enumerate}

\item \textbf{Agent-Aware Reachability in Differentiable Layout Optimization}: The LayoutVLM framework is extended by integrating explicit reachability constraints for embodied agents, such as robots, humans, or animals, directly into the differentiable optimization process. This allows the model to generate 3D layouts that are not only semantically coherent but also physically navigable and actionable by the intended agent.

\item \textbf{Efficient Local Refinement for Improved Layout Stability}: A local refinement stage is introduced that selectively re-optimizes only problematic object placements while keeping the rest of the scene fixed. This improves convergence efficiency, ensures higher-quality layouts, and maintains semantic and structural consistency without increasing global optimization iterations.

\end{enumerate}

\section{Related Work}

\subsection{LLM and VLM-Based Layout Generation}

Large language models (LLMs) and vision--language models (VLMs) have shown strong capabilities in understanding complex relationships and generating structured outputs that encode both visual and semantic information. LayoutGPT guides LLMs through prompts and templates to generate object parameters for scene layouts \hyperref[sec:references]{[16]}. Graph-based scene generation approaches treat objects as nodes and synthesize scenes using learned relational structures \hyperref[sec:references]{[17]}. 

Agentic and iterative frameworks have also been explored for 3D layout generation. SceneWeaver employs a reflective agentic approach that integrates language-model planning with generation tools to produce physically plausible indoor scenes \hyperref[sec:references]{[20]}. Holodeck uses multi-round interactions with LLMs to generate semantically consistent 3D environments, including floors, walls, and object placements \hyperref[sec:references]{[21]}. SceneX adopts a multi-agent formulation to construct explicit world models and generate executable layout scripts \hyperref[sec:references]{[22]}.

While purely LLM-based methods achieve impressive results, they often struggle with limited layout diversity, rendering artifacts, and weak global planning, particularly in complex scenes.

\subsection{Optimization for 3D Scene Generation}

Constraint-based and optimization-driven approaches have long been used in 3D scene generation to balance multiple objectives derived from physical laws, functional requirements, and spatial design principles. Physical constraints include object interpenetration, stability, and support, while layout-level constraints capture accessibility and functional relationships \hyperref[sec:references]{[11]} \hyperref[sec:references]{[12]}.

Data-driven models learn probabilistic object arrangements or hierarchical scene structures using graphical or neural representations \hyperref[sec:references]{[13]} \hyperref[sec:references]{[14]}. Hybrid discrete--continuous optimization techniques are particularly effective in settings that combine categorical decisions with continuous spatial parameters. LayoutVLM integrates vision--language supervision with differentiable optimization to refine 3D layouts from natural language descriptions. It improves spatial planning through complementary representations and self-consistent decoding \hyperref[sec:references]{[1]} this is promising especially for use-cases where control and explainability are indispensable.

 \section{Architecture}
 
\begin{figure*}[tp]
  \centering
  \includegraphics[width=0.75\textwidth]{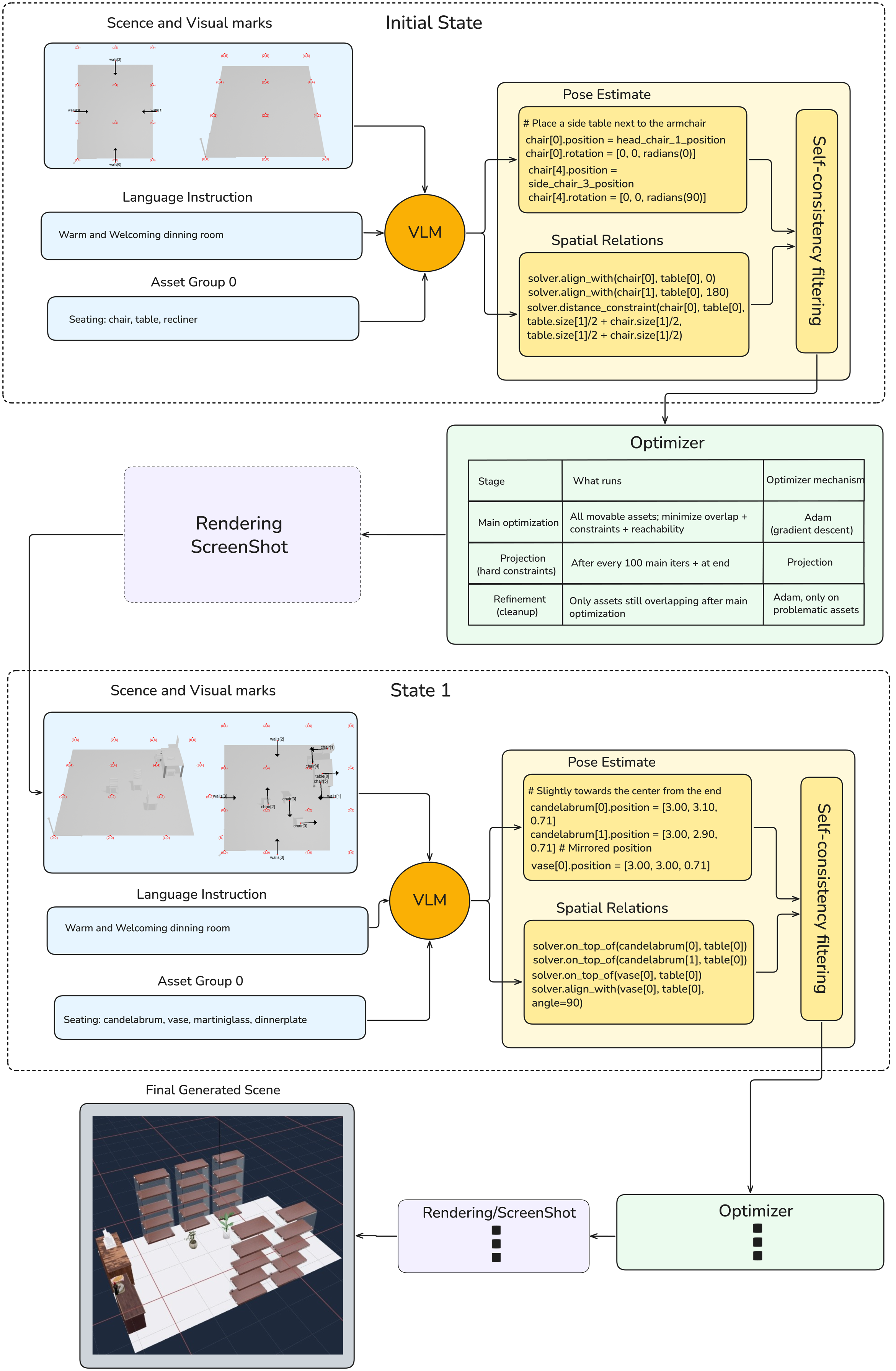}
  \caption{Overall architecture of RoboLayout}
  \label{fig:architecture}
\end{figure*}

RoboLayout comprises three main layers. Orchestration: The central orchestrator responsible for coordinating groupings, rendering and other cognitive processes of the solver and sandbox.
Sandbox: Translates constraints into feasible scene layouts. Solver: optimizer based on hard and soft constraints 
for spatial arrangements and refinement of the final optimized scene. As shown in Figure ~\ref{fig:architecture}, the layered architecture of RoboLayout is composed of following steps:

\paragraph{Initial state.}
The pipeline takes the language instruction (task\_description, layout\_criteria) 
and the room shape (boundary.floor\_vertices, wall\_height) plus the list of assets 
as the initial program state. The sandbox is initialized with walls and asset variables; positions and rotations 
are set to random feasible placements inside the boundary.

\paragraph{Furniture grouping}
An LLM groups furniture based on the prompt and task 
(e.g.\ ``bed + nightstands'', ``rug'', ``seating''). 
Each group has a name and key spatial relations between assets, placement then runs group by group with group-specific layout criteria.

\paragraph{Pose estimation and spatial relations.}
For each (group of) assets, the system produces pose and spatial relations
by calling an LLM with the current scene (top-down and side renderings) 
and the layout criteria. The LLM returns a constraint program: high-level relations such as 
against\_wall, align\_with, distance\_constraint.

\paragraph{Conversion to Python program.}
The LLM output is parsed into executable Python: constraint calls (e.g.\ solver.against\_wall(...), solver.distance\_constraint(...)) 
are executed in the sandbox. That updates the in-memory constraint list.
\paragraph{Optimization with reachability.}
The gradient-based solver (e.g.\ GradSolver.optimize) 
minimizes a loss over positions and rotations: overlap loss (no intersection), 
existing- and new-constraint losses (satisfy spatial relations),
and reachability loss. Reachability encourages clearance between furniture so a virtual disc of radius 
robot\_radius can pass; it is disabled if robot\_radius is None or $\le 0$. Optimization runs per group (or once in one\_shot), producing poses and optional per-step GIFs.

\paragraph{Refinement.}
After the main optimization, a cleanup step identifies problematic pairs (e.g.\ overlapping footprints), 
freezes non-problematic assets, and re-runs optimization only for the problematic subset.
This refines the answer without full-scene re-optimization.

\subsubsection{Self-Consistency Filter}
Self-consistent decoding typically refers to sampling
multiple outputs from the model (e.g.\ several constraint programs), 
 then selecting one by a criterion (e.g.\ majority vote or best score). 
 This codebase does not use self-consistent decoding: 
 it obtains a single LLM constraint program per group (with retries on parse/execution failure), 
 not multiple samples followed by selection. Self-consistency filter is what is implemented: before optimization, the sandbox 
runs self\_consistency\_filtering 
on the new constraints. It checks consistency with the current scene and existing constraints: 
rejects duplicate or conflicting constraints 
(e.g.\ duplicate against\_wall, or a second orientation constraint on the same object), 
resolves against\_wall to the nearest wall, and tightens distance constraints to feasible 
ranges.

\subsection{Orchestration}

This module coordinates the layout generation pipeline through semantic grouping, visual reasoning, and constraint synthesis.  
The task is formulated as an optimization problem
\[
\mathcal{P}_{\text{layout}} = (\mathcal{F}, \mathcal{O}, \mathcal{D}),
\]
where $\mathcal{F}$ is the set of furniture items, $\mathcal{O}$ is the layout objective, and $\mathcal{D}$ describes the room domain.

\subsubsection{\textbf{Furniture Grouping}}

A language-driven grouping function
\[
\mathcal{G}: \mathcal{O} \rightarrow \{G_1, \dots, G_K\}
\]
maps the objective to an ordered list of furniture groups. Each group is defined as
\[
G_k = (g_k, \mathcal{F}_k, \mathcal{R}_k),
\]
where
\begin{align}
g_k &:\ \text{semantic group label and rationale}, \\
\mathcal{F}_k &\subseteq \mathcal{F}:\ \text{furniture items in group } k, \\
\mathcal{R}_k &:\ \text{intra-group spatial relationships}.
\end{align}
Groups are processed from structural to decorative elements:
\[
\text{priority}(G_1) < \text{priority}(G_2) < \dots < \text{priority}(G_K).
\]

\textit{Constraint Generation.}  
At step $t$, the current scene state is
\[
\mathcal{S}_t = \{(\mathbf{p}_i, \boldsymbol{\theta}_i) \mid f_i \in \mathcal{F}_{\text{placed}}\},
\]
where $\mathbf{p}_i \in \mathbb{R}^3$ and $\boldsymbol{\theta}_i$ denote position and orientation.  
Rendered visual observations are grouped as
\[
\mathcal{I}_t = \{I_{\text{top}}, I_{\text{side}}, I_{f_1}, \dots, I_{f_{|\mathcal{F}_k|}}\}.
\]
A vision--language model then produces executable constraints:
\[
\mathcal{C}_k = \mathrm{VLM}(\mathcal{S}_t, \mathcal{F}_k, \mathcal{D}, \mathcal{I}_t).
\]

These constraints are represented as a set of parametric functions over furniture poses:
\[
\mathcal{C}_k = \left\{ 
\phi_j(\mathbf{p}_{a_j}, \boldsymbol{\theta}_{a_j}, \mathbf{p}_{b_j}, \boldsymbol{\theta}_{b_j}; \,\boldsymbol{\lambda}_j) 
\right\}_{j=1}^{J_k},
\]
where each $\phi_j$ converts a qualitative spatial rule into a quantitative geometric relation.

\textit{Iterative Positioning.}  
Furniture is placed group by group using constrained optimization:
\begin{align}
\mathcal{F}_{\text{placed}}^{(0)} &= \varnothing, \\
\mathcal{F}_{\text{placed}}^{(k)} &= \mathcal{F}_{\text{placed}}^{(k-1)} 
\cup 
\operatorname*{arg\,opt}_{\{\mathbf{p}_i,\boldsymbol{\theta}_i\}_{f_i \in \mathcal{F}_k}}
\mathcal{C}_k,
\end{align}
for $k = 1,\dots,K$, until all items are positioned.

\subsubsection{\textbf{Rendering}}

This module renders 3D scenes and projects world coordinates into image space for visual feedback.

\textit{Camera Placement.}  
Let $\mathbf{B}_x, \mathbf{B}_y$ denote all furniture bounding coordinates in the floor plane. The room center and scale are
\begin{align}
\mathbf{c}_{\text{floor}} &= 
\begin{bmatrix}
\frac{\max(\mathbf{B}_x)+\min(\mathbf{B}_x)}{2} \\
\frac{\max(\mathbf{B}_y)+\min(\mathbf{B}_y)}{2} \\
0
\end{bmatrix}, \\
W_{\text{room}} &= 
\max\!\big(\max(\mathbf{B}_x)-\min(\mathbf{B}_x),\;
           \max(\mathbf{B}_y)-\min(\mathbf{B}_y)\big).
\end{align}

Camera height is scaled proportionally:
\[
H_{\text{cam}} = \alpha \, W_{\text{room}}, \quad \alpha = 1.5.
\]

A general camera position using spherical angles $(\phi,\theta)$ is
\[
\mathbf{c}_{\text{cam}} = \mathbf{c}_{\text{floor}} +
H_{\text{cam}}
\begin{bmatrix}
\sin\phi \cos\theta \\
\sin\phi \sin\theta \\
\cos\phi
\end{bmatrix}.
\]

\textit{Coordinate Grid Overlay.}  
Grid spacing adapts to room size:
\[
\Delta = \max\!\left(\Delta_0,\; \left\lceil \frac{W_{\text{room}}}{500} \right\rceil \right).
\]
Grid points on the floor are
\[
(x_i, y_j) = (x_{\min}, y_{\min}) + (i\Delta, j\Delta),
\]
and their image projections form the overlay set
\[
\mathcal{M} = \left\{ (u_{ij}, v_{ij}) = 
\operatorname{project}\!\left([x_i, y_j, 0]^T\right) \right\}.
\]

\textit{Furniture Orientation Annotation.}  
For furniture $f_i$ with yaw angle $\theta_{i,z}$,
\[
\mathbf{p}_{i}^{\text{dir}} = 
\mathbf{p}_i + \ell
\begin{bmatrix}
\cos\theta_{i,z} \\
\sin\theta_{i,z} \\
0
\end{bmatrix}, 
\quad \ell = 0.75.
\]
The annotation in image space is
\[
\big( \operatorname{project}(\mathbf{p}_i),\;
       \operatorname{project}(\mathbf{p}_{i}^{\text{dir}}),\;
       \text{label}_i \big).
\]

\textit{Wall Annotation.}  
For a wall segment $(\mathbf{v}_i, \mathbf{v}_{i+1})$, let
\[
\mathbf{d}_i = \mathbf{v}_{i+1} - \mathbf{v}_i, 
\quad 
\hat{\mathbf{n}}_i = 
\frac{\mathbf{R}_{90^\circ}\mathbf{d}_i}{\|\mathbf{d}_i\|_2},
\]
where $\mathbf{R}_{90^\circ}$ rotates a 2D vector by $90^\circ$.  
The wall label position is then
\[
\mathbf{c}_{\text{wall}} = 
\frac{\mathbf{v}_i + \mathbf{v}_{i+1}}{2} - \beta \hat{\mathbf{n}}_i,
\quad \beta = 0.2.
\]

\subsection{\textbf{Sandbox}}

The sandbox module converts high-level constraint descriptions produced by the VLM into executable optimization code. It maintains a local namespace in which language-generated Python snippets map abstract rules 
(e.g., ``furniture A should be near wall B'') into concrete geometric constraint functions.  
Also it filters invalid constraints, resolves conflicts, and ensures all constraints correspond to feasible spatial relationships before optimization.  
Iterative multi-objective optimization is employed with soft and hard constraints, where hard constraints enforce non-negotiable often physical constraints.

\subsubsection{\textbf{Out-of-Bounds Placement Constraint:} }
For each furniture item $f_i$ with position $\mathbf{p}_i = [x_i, y_i, z_i]^T$, orientation $\boldsymbol{\theta}_i$, and size $\mathbf{s}_i$, its oriented 2D footprint is computed as
\[
\mathbf{C}_i = \operatorname{polygon}(\mathbf{p}_i, \boldsymbol{\theta}_i, \mathbf{s}_i),
\]
where $\mathbf{C}_i \in \mathbb{R}^{4 \times 2}$ contains the four ground-plane corners.  
The room boundary is a polygon $\mathcal{P}$ with counter-clockwise vertices $\{\mathbf{b}_1,\dots,\mathbf{b}_n\}$.  
A tolerance-expanded region is defined as $\mathcal{P}_\epsilon = \mathcal{P} \oplus B(0,\epsilon)$.

If any corner $\mathbf{c}_{i,k}$ lies outside $\mathcal{P}_\epsilon$, a corrective translation $\boldsymbol{\tau}_i$ is computed.

\textbf{Case 1: All corners outside.}
\begin{align}
\mathbf{p}_i^{2D} &= [x_i, y_i]^T \\
\mathbf{p}_{\text{proj}} &= \arg\min_{\mathbf{p}\in\partial\mathcal{P}} \|\mathbf{p}-\mathbf{p}_i^{2D}\|_2 \\
\hat{\mathbf{v}} &= \frac{\mathbf{c}_{\mathcal{P}}-\mathbf{p}_{\text{proj}}}{\|\mathbf{c}_{\mathcal{P}}-\mathbf{p}_{\text{proj}}\|_2} \\
\boldsymbol{\tau}_i &= \hat{\mathbf{v}}\big(\|\mathbf{p}_{\text{proj}}-\mathbf{p}_i^{2D}\|_2 + \epsilon + \delta\big)
\end{align}
where $\mathbf{c}_{\mathcal{P}}$ is the room centroid and $\delta=0.15$ m is a safety margin.

\textbf{Case 2: Partial violation.}
\[
\boldsymbol{\tau}_i = \frac{1}{|\mathcal{K}_{\text{out}}|}\sum_{k\in\mathcal{K}_{\text{out}}} \boldsymbol{\tau}_{i,k},
\]
where $\mathcal{K}_{\text{out}}$ indexes corners outside $\mathcal{P}_\epsilon$.

The position update is
\[
\mathbf{p}_i^{[t+1]}[:2] = \mathbf{p}_i^{[t]}[:2] + \boldsymbol{\tau}_i,
\]
iterated up to $M=10$ times or until all corners satisfy $\mathbf{C}_i \subseteq \mathcal{P}_\epsilon$.

\begin{algorithmic}[1]
\REQUIRE Furniture $\mathcal{F}$, boundary $\mathcal{P}$, tolerance $\epsilon$, margin $\delta$, max steps $M$
\FOR{each $f_i \in \mathcal{F}$}
    \STATE $t \gets 0$
    \WHILE{$t < M$ and $\mathbf{C}_i \not\subseteq \mathcal{P}_\epsilon$}
        \STATE Compute $\boldsymbol{\tau}_i$ based on violation case
        \STATE $\mathbf{p}_i[:2] \gets \mathbf{p}_i[:2] + \boldsymbol{\tau}_i$
        \STATE Recompute $\mathbf{C}_i$
        \STATE $t \gets t+1$
    \ENDWHILE
\ENDFOR
\end{algorithmic}

\subsubsection{\textbf{Rotation Constraint}}
For stabilizing stochastic rotations during optimization updates, each proposed step is projected onto the feasible manifold before acceptance.  
Given the current state $\mathbf{x}$ and a candidate step $\mathbf{s}$ (e.g., from Adam), the following conditions are enforced:
\[
\|\mathbf{s}\| \le \Delta, 
\quad \mathbf{x} + \mathbf{s} \in \mathcal{B}, 
\quad c_{\text{rot}}(\mathbf{x} + \mathbf{s}) = 0,
\]
where $\Delta$ is the trust-region radius, $\mathcal{B}$ denotes box bounds on positions and rotations, and $c_{\text{rot}}$ encodes rotation feasibility.  
This guarantees that all accepted updates remain inside the feasible set. Each rotation angle $\theta_i$ is parameterized using $(\cos\theta_i,\sin\theta_i)$ constrained to lie on the unit circle:
\[
c_{\text{rot}}(\mathbf{x}) = \cos^2\theta_i + \sin^2\theta_i - 1 = 0
\quad \forall i.
\]

The constrained optimization problem is therefore
\[
\min_{\mathbf{x} \in \mathcal{B}} \; f(\mathbf{x})
\quad \text{s.t.} \quad c_{\text{rot}}(\mathbf{x}) = 0.
\]

Convergence tolerances are
\[
\text{gtol} = \text{xtol} = \text{barrier\_tol} = 10^{-12}.
\]

\begin{algorithmic}[1]
\REQUIRE Initial state $\mathbf{x}_0$, objective $f$, bounds $\mathcal{B}$
\STATE $\mathbf{x} \gets \mathbf{x}_0$, \quad $\Delta \gets \Delta_0$
\WHILE{not converged}
    \STATE Propose step $\mathbf{s}$
    \STATE Project $\mathbf{s}$ so that $\|\mathbf{s}\|\le\Delta$, $\mathbf{x}+\mathbf{s}\in\mathcal{B}$, $c_{\text{rot}}(\mathbf{x}+\mathbf{s})=0$
    \STATE Compute ratio $\rho = \dfrac{f(\mathbf{x}) - f(\mathbf{x}+\mathbf{s})}{\text{predicted decrease}}$
    \IF{$\rho > \eta_1$}
        \STATE $\mathbf{x} \gets \mathbf{x} + \mathbf{s}$
        \IF{$\rho > \eta_2$}
            \STATE $\Delta \gets \min(\gamma_2 \Delta, \Delta_{\max})$
        \ENDIF
    \ELSE
        \STATE $\Delta \gets \gamma_1 \Delta$
    \ENDIF
\ENDWHILE
\RETURN $\mathbf{x}$
\end{algorithmic}

\subsubsection{\textbf{Reachability Constraint}}
The reachability constraint is implemented as a \emph{walking robot governed by pairwise clearance rule}. For every pair of movable (non--on-top-of) assets $i$, $j$, It takes their center positions $p_i$, $p_j$ in the ground plane and requires the center--center distance to be at least $e_i + e_j + 2r$ (half-footprint of $i$ plus half-footprint of $j$ plus robot diameter), then penalizes (via a soft loss) when the actual distance is smaller than this required clearance.

Working in a top--down 2D view of the room. Let
\begin{itemize}
  \item $\mathcal{A}$ be the set of all movable, non--fixture, non--ceiling assets that are optimized by the gradient solver.
  \item For each asset $i \in \mathcal{A}$, let its center position in the ground plane be
  \[
    p_i \in \mathbb{R}^2,
  \]
  corresponding to the $(x,y)$ components of its 3D position.
  \item Let its axis-aligned bounding box dimensions in the ground plane be $w_i$ (width in $x$) and $h_i$ (height in $y$). An effective in-plane ``radius'' is defined as
  \[
    e_i \triangleq \frac{1}{2} \min(w_i, h_i),
  \]
  so that the footprint is approximated by a disc of radius $e_i$.
  \item Let $r > 0$ be the radius of the virtual robot which is provided via \verb|--robot_radius| through code intialization.
\end{itemize}

Intuitively, the objective is for a robot of radius $r$ to be able to move between any two assets $i$ and $j$ without colliding with their footprints. In a simple disc approximation, this means the distance between the centers $p_i$ and $p_j$ should be at least
\[
  e_i + e_j + 2r.
\]

For each unordered pair $(i,j)$ of assets with $i \neq j$, a required center-to-center distance is defined as
\begin{equation}
  d_{ij}^{\mathrm{req}} \triangleq e_i + e_j + 2r.
\end{equation}

Let
\begin{equation}
  d_{ij} \triangleq \lVert p_i - p_j \rVert_2
\end{equation}
be the actual Euclidean distance between the asset centers in the ground plane.

Ideally, the goal is to enforce the hard constraint
\begin{equation}
  d_{ij} \ge d_{ij}^{\mathrm{req}} \quad \text{for all pairs } (i,j).
\end{equation}
Instead of imposing this as a strict feasibility constraint, it is relaxed into a differentiable soft penalty that can be optimized by gradient descent.

In the implementation, squared distances are used to keep the loss smooth and easy to differentiate. For each pair $(i,j)$, the squared distance is considered
\begin{equation}
  D_{ij} \triangleq \lVert p_i - p_j \rVert_2^2,
\end{equation}
and the squared required distance
\begin{equation}
  R_{ij} \triangleq \bigl(d_{ij}^{\mathrm{req}}\bigr)^2
  = \bigl(e_i + e_j + 2r\bigr)^2.
\end{equation}

The pairwise violation is then defined as
\begin{equation}
  v_{ij} \triangleq \max\bigl(0,\; R_{ij} - D_{ij}\bigr).
\end{equation}
This quantity is zero whenever the actual squared distance $D_{ij}$ is at least the required squared distance $R_{ij}$, and grows linearly with the shortfall when assets are too close.

The unscaled \emph{reachability loss} over all eligible pairs is
\begin{equation}
  L_{\mathrm{reach}}^{\mathrm{raw}}
  \triangleq
  \sum_{(i,j) \in \mathcal{P}} v_{ij},
\end{equation}
where $\mathcal{P}$ is the set of unordered asset pairs that:
\begin{itemize}
  \item are both members of $\mathcal{A}$ (movable, non--fixture, non--ceiling), and
  \item are not marked as ``on top of'' one another (those pairs are handled by a separate constraint).
\end{itemize}

To keep the magnitude of this loss term comparable to other components (overlap, existing constraints, new constraints), a small scalar factor $\alpha > 0$ is applied:
\begin{equation}
  L_{\mathrm{reach}} \triangleq
  \alpha \; L_{\mathrm{reach}}^{\mathrm{raw}}
  = \alpha \sum_{(i,j) \in \mathcal{P}} \max\bigl(0, R_{ij} - D_{ij}\bigr).
\end{equation}

In the current implementation, $\alpha = 0.01$.

The reachability loss is combined with the other layout losses in the gradient solver. Let
\begin{itemize}
  \item $L_{\mathrm{overlap}}$ be the non-overlap (bounding-box) loss,
  \item $L_{\mathrm{exist}}$ be the loss from existing constraints,
  \item $L_{\mathrm{new}}$ be the loss from new constraints for the current group of assets,
  \item $L_{\mathrm{reach}}$ be the reachability loss defined above.
\end{itemize}

The total objective minimized by gradient descent is
\begin{equation}
  L_{\mathrm{total}} \;=\;
  L_{\mathrm{overlap}}
  + L_{\mathrm{exist}}
  + L_{\mathrm{new}}
  + L_{\mathrm{reach}}.
\end{equation}

When $r \le 0$ or no robot radius is provided, the implementation sets $L_{\mathrm{reach}} = 0$, effectively disabling the reachability term.

Table~\ref{tab:constraints} shows the list of the hard and soft constraints used by the solver.
Hard constraints are enforced by projection or by constraints in the solver. 
Soft constraints are implemented as differentiable loss terms that are minimized and are not strictly enforced.

\begin{table*}[tp]
  \centering
  \caption{List of hard and soft constraints.}
  \label{tab:constraints}
  \begin{tabular}{@{}llp{0.5\textwidth}@{}}
    \toprule
    Type & Name & Rule \\
    \midrule
    Hard & Boundary (inside room) & Every asset's 2D footprint must lie inside the room polygon. Any corner outside is projected onto the boundary. \\
    Hard & On-top (vertical + horizontal) & For each on\_top\_of(A,B): (1) $A$'s $z$ is set so $A$ sits on top of $B$ ($z = B.z + B.h/2 + A.h/2$). (2) $A$'s $xy$ is projected onto $B$'s 2D polygon so $A$ is centered over $B$. \\
    Hard & Rotation unit norm & When using solver\_type of minimize, rotation is parameterized as $(\cos\theta, \sin\theta)$ with $\cos^2\theta + \sin^2\theta = 1$ via NonlinearConstraint. \\
    \midrule
    Soft & Overlap (no intersection) & 2D footprints of non--on-top pairs must not overlap. Loss: IoU / bbox overlap (e.g.\ scaled by 1000); minimized so overlap goes to zero. \\
    Soft & Reachability & Pairs of objects should leave enough clearance for a disc of radius robot\_radius. Loss: penalize if center distance $< (r_a + r_b + 2\times\text{robot\_radius})$. \\
    Soft & Against\_wall & Asset should be (1) close to a wall segment and (2) front parallel to the wall. Loss: distance of footprint to wall + angle between asset front and wall direction. \\
    Soft & Distance\_constraint & Distance between two centers should be in $[\mathit{min\_distance}, \mathit{max\_distance}]$. Loss: penalize when outside the range (scaled by weight). \\
    Soft & Point\_towards & Asset1's front should point toward asset2. Loss: 0 if the ray from asset1 in its front direction hits asset2's polygon; otherwise cosine alignment loss with the direction to asset2. \\
    Soft & Align\_with & Two assets' forward directions should be aligned (optional angle offset). Loss: cosine distance between (possibly rotated) forward vectors. \\
    Soft & On\_top\_of (soft part) & In addition to the hard on-top placement, the upper object should stay over the lower in 2D. Loss: negative IoU (clamped) so that overlap in the horizontal plane is encouraged. \\
    Soft & Symmetric\_pair & Two assets should be mirror-symmetric about a reference (wall or central asset). Loss: position reflection error + optional orientation mirror alignment. \\
    \bottomrule
  \end{tabular}
\end{table*}
\noindent

\subsection{\textbf{Solver}}

Layout generation problem is solved as a hybrid discrete–continuous multi-objective optimization problem. Hard constraints are enforced via projection steps that keep every intermediate state feasible, while soft constraints remain in the differentiable objective and are optimized with gradient-based updates.

\subsubsection{\textbf{Objective Function}}
Let each soft constraint $\phi$ act on a subset of furniture indices $\mathcal{I}$. The total loss is
\[
\mathcal{L}_{\text{total}} = \sum_{(\phi,\mathcal{I})} \phi(\mathbf{p}_{\mathcal{I}}, \boldsymbol{\theta}_{\mathcal{I}}),
\qquad
\mathcal{L} = 0.1\,\mathcal{L}_{\text{total}}.
\]

\subsubsection{\textbf{Learning Rate Scaling}}
To normalize updates across room sizes, the following quantity is defined:
\[
R_{\text{room}} = \max\big(\max(\mathbf{B}_x)-\min(\mathbf{B}_x),\;
\max(\mathbf{B}_y)-\min(\mathbf{B}_y),\;1\big),
\]
and use separate learning rates
\[
\alpha_{\text{pos}} = \frac{3\alpha_0}{R_{\text{room}}}, 
\qquad
\alpha_{\text{rot}} = \alpha_0,
\]
where $\alpha_0$ is the base rate (typically $0.03$).

\textit{Adam Update.}  
For any parameter $\boldsymbol{\xi}\in\{\mathbf{p}_i,\boldsymbol{\theta}_i\}$,
\begin{align}
m_t &= \beta_1 m_{t-1} + (1-\beta_1)\nabla_{\boldsymbol{\xi}}\mathcal{L}, \\
v_t &= \beta_2 v_{t-1} + (1-\beta_2)(\nabla_{\boldsymbol{\xi}}\mathcal{L})^2, \\
\boldsymbol{\xi}_t &= \boldsymbol{\xi}_{t-1} - \alpha_t \frac{m_t/(1-\beta_1^t)}{\sqrt{v_t/(1-\beta_2^t)} + \epsilon},
\end{align}
with $\beta_1=\beta_2=0.9$ and $\epsilon=10^{-8}$.  
Gradients are clipped to $\|\nabla_{\boldsymbol{\xi}}\mathcal{L}\|_2 \le 10$.

The learning rate decays every 100 steps:
\[
\alpha_t = \alpha \cdot \gamma^{\lfloor t/100 \rfloor}, 
\qquad \gamma = 0.95.
\]

\subsubsection{\textbf{Feasibility Enforcement}}
After each update, parameters are projected onto the feasible manifold (bounds + rotation constraints).  
Additional hard corrections are applied periodically, for boundary projection every 80 steps.

\begin{algorithmic}[1]
\REQUIRE Furniture $\mathcal{F}$, iterations $T$, base rate $\alpha_0$
\STATE Initialize Adam moments $m_0,v_0 \gets 0$
\STATE Compute $R_{\text{room}}$, $\alpha_{\text{pos}}$, $\alpha_{\text{rot}}$
\FOR{$t=1$ to $T$}
    \STATE Compute $\mathcal{L}$ and gradients
    \STATE Clip gradients
    \STATE Update parameters using Adam
    \STATE Project parameters onto feasible manifold
    \IF{$t \bmod 100 = 0$} \STATE Enforce boundary constraints \ENDIF
\ENDFOR
\end{algorithmic}

\noindent
Applying a trust-region style projection at every iteration filters noisy gradient steps.  
Well-aligned updates pass through unchanged, while unstable steps are reduced or rejected, leading to smoother convergence without sacrificing the exploratory benefits of Adam.

\subsubsection{\textbf{Refinement}}
Running the main layout optimization until full convergence often requires a large number of iterations and is costly. The \emph{refinement} (or \emph{cleanup}) step avoids this by shrinking the optimization space: after the main run, only the \emph{problematic} assets (e.g.\ those still overlapping) are identified and \emph{just those} are re-optimized while keeping the rest fixed. Thus there is no need to increase the main iteration count to converge; instead, a short, cheaper pass over a smaller variable set resolves remaining conflicts in less time. In the following, the terms ``refinement'' or ``cleanup'' are used for this step (the code also refers to it as cleanup). 

After the main optimization, \emph{problematic} assets are identified as those involved in at least one overlapping pair (excluding fixtures and stacked pairs):
\begin{equation}
  \mathcal{I}_{\text{prob}} = \bigl\{ i \in \mathcal{I} \setminus \mathcal{F} \;\big|\; \exists j \ne i:\; (i,j) \notin \mathcal{S},\; \mathcal{P}_i \cap \mathcal{P}_j \ne \emptyset \bigr\}.
\end{equation}
In code, polygon intersection is tested with Shapely (e.g.\ \texttt{poly\_i.intersects(poly\_j)}). Movable assets are partitioned into:
\begin{itemize}
  \item \textbf{Correct (frozen):} $\mathcal{I}_{\text{fix}} = (\mathcal{I} \setminus \mathcal{F}) \setminus \mathcal{I}_{\text{prob}}$. Their positions and rotations are fixed at the current values $\bar{\mathbf{p}}_i$, $\bar{\mathbf{r}}_i$.
  \item \textbf{Problematic (optimized):} $\mathcal{I}_{\text{prob}}$. Only these variables are updated during cleanup.
\end{itemize}

Let $\mathbf{p}_{\mathcal{I}_{\text{prob}}}$ and $\mathbf{r}_{\mathcal{I}_{\text{prob}}}$ denote the positions and rotations of problematic assets; all others are fixed. The cleanup step solves, for a small number of iterations $T_{\text{cleanup}}$ (e.g.\ $40$), the same total loss but with gradients taken only with respect to $\mathbf{p}_i$, $\mathbf{r}_i$ for $i \in \mathcal{I}_{\text{prob}}$:
\begin{equation}
  \min_{\mathbf{p}_i,\mathbf{r}_i,\, i \in \mathcal{I}_{\text{prob}}} \mathcal{L}_{\text{total}}\bigl( \mathbf{p}_{\mathcal{I}_{\text{prob}}}, \mathbf{r}_{\mathcal{I}_{\text{prob}}};\; \bar{\mathbf{p}}_{\mathcal{I}_{\text{fix}}}, \bar{\mathbf{r}}_{\mathcal{I}_{\text{fix}}} \bigr).
\end{equation}

Only parameters for $i \in \mathcal{I}_{\text{prob}}$ have \texttt{requires\_grad=True}; others are frozen, so constraint terms that depend only on fixed assets do not require gradients (handled by \texttt{allow\_nograd\_constraints}). Optimizer: Adam with learning rate $\eta_{\text{cleanup}}$ (e.g.\ $0.01$), gradient clipping (max norm $1$). Every $K$ steps (e.g.\ $10$), apply the projection $\Pi$ to keep layouts feasible.

\begin{algorithmic}[1]
\REQUIRE Furniture set $\mathcal{F}$, problematic furniture $\mathcal{F}_{\text{problem}}$, iterations $T_{\text{ft}}$, learning rate $\alpha_{\text{ft}}$
\FOR{each furniture $f_i \in \mathcal{F}$}
    \IF{$f_i \notin \mathcal{F}_{\text{problem}}$}
        \STATE $f_i.\text{optimize} \gets 0$
        \STATE $\nabla_{\mathbf{p}_i} \mathcal{L} \gets 0$, $\nabla_{\boldsymbol{\theta}_i} \mathcal{L} \gets 0$
    \ELSE
        \STATE $f_i.\text{optimize} \gets 1$
        \STATE $\mathbf{p}_i.\text{requires\_grad} \gets \text{True}$
        \STATE $\boldsymbol{\theta}_i.\text{requires\_grad} \gets \text{True}$
    \ENDIF
\ENDFOR
\STATE $\text{allow\_nograd\_constraints} \gets \text{True}$
\FOR{$t = 1$ to $T_{\text{ft}}$}
    \STATE Compute $\nabla_{\mathbf{p}_i} \mathcal{L}$, $\nabla_{\boldsymbol{\theta}_i} \mathcal{L}$ for $f_i \in \mathcal{F}_{\text{problem}}$
    \STATE Update $\mathbf{p}_i$, $\boldsymbol{\theta}_i$ using Adam with learning rate $\alpha_{\text{ft}}$
\ENDFOR
\end{algorithmic}

\section{Experiments}

\subsection{Experimental Setup}

Evaluation of \textit{RoboLayout} on the task of 3D indoor scene layout generation is done under the following configuration:
\begin{itemize}
    \item VLM: GPT-4o
    \item Optimization Strategy: Iterative multi-objective optimization
    \item Spatial Configurations: Different rooms with varying shapes and dimensions
    \item Object Sets: 5--10 furniture and decorative elements per scene
    \item Contextual Constraints: Structural realism and semantic placement consistency
\end{itemize}

\subsection{Qualitative Results}

\begin{figure*}[tp]
  \vspace*{0pt}
  \centering
  \begin{subfigure}[b]{0.24\linewidth}
      \centering
      \includegraphics[width=\textwidth]{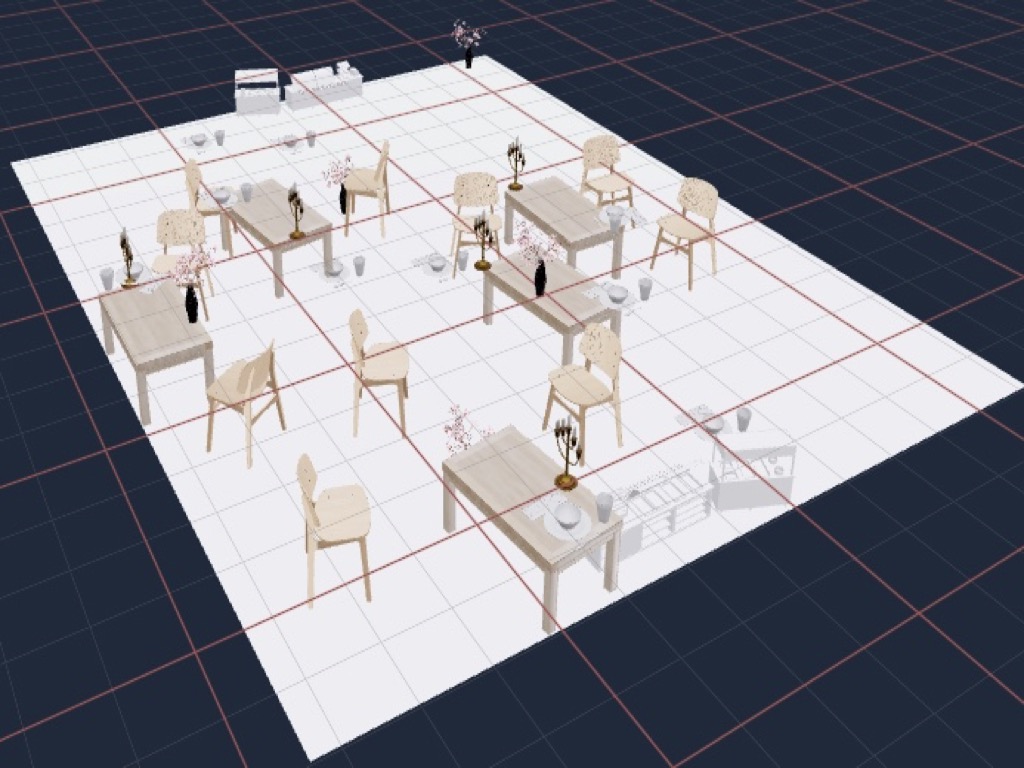}
      \caption{buffet restaurant}
      \label{fig:gl1}
  \end{subfigure}
  \hfill
  \begin{subfigure}[b]{0.24\linewidth}
      \centering
      \includegraphics[width=\textwidth]{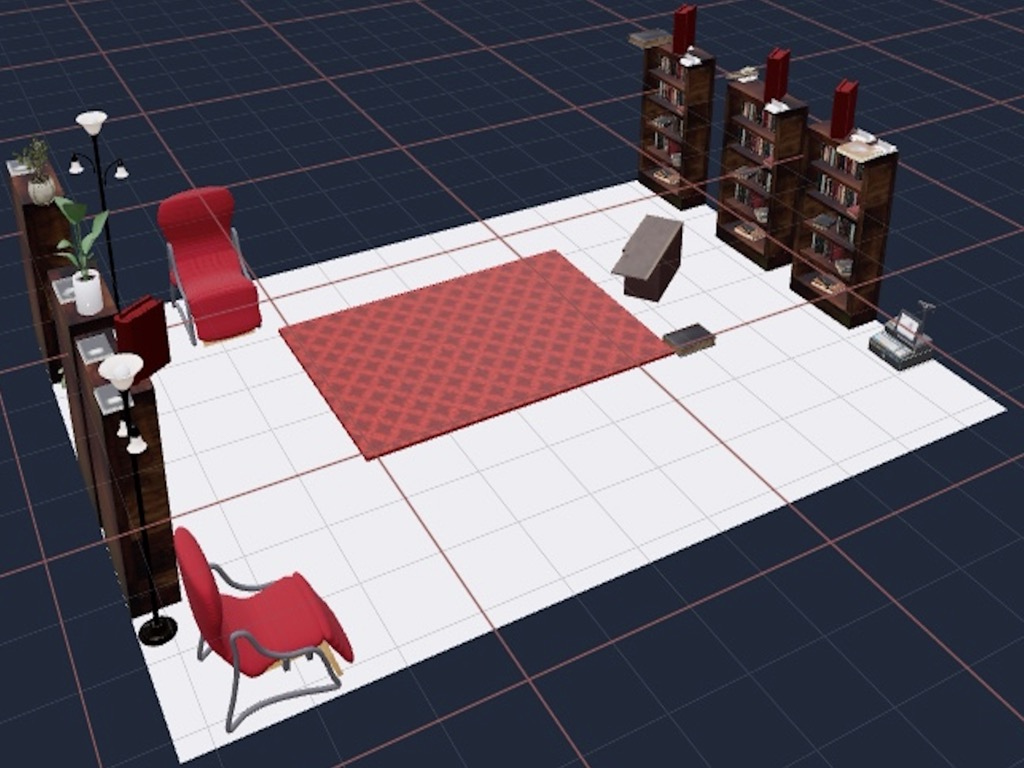}
      \caption{bookstore}
      \label{fig:gl2}
  \end{subfigure}
  \hfill
  \begin{subfigure}[b]{0.24\linewidth}
      \centering
      \includegraphics[width=\textwidth]{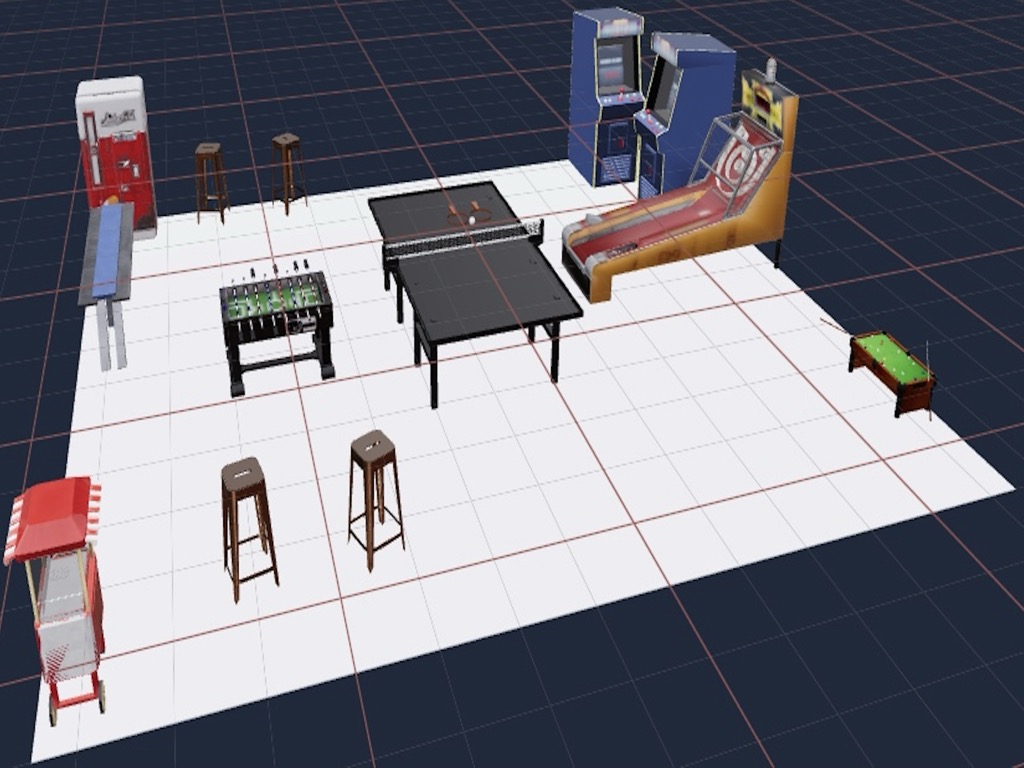}
      \caption{game room}
      \label{fig:gl3}
  \end{subfigure}
  \hfill
  \begin{subfigure}[b]{0.24\linewidth}
      \centering
      \includegraphics[width=\textwidth]{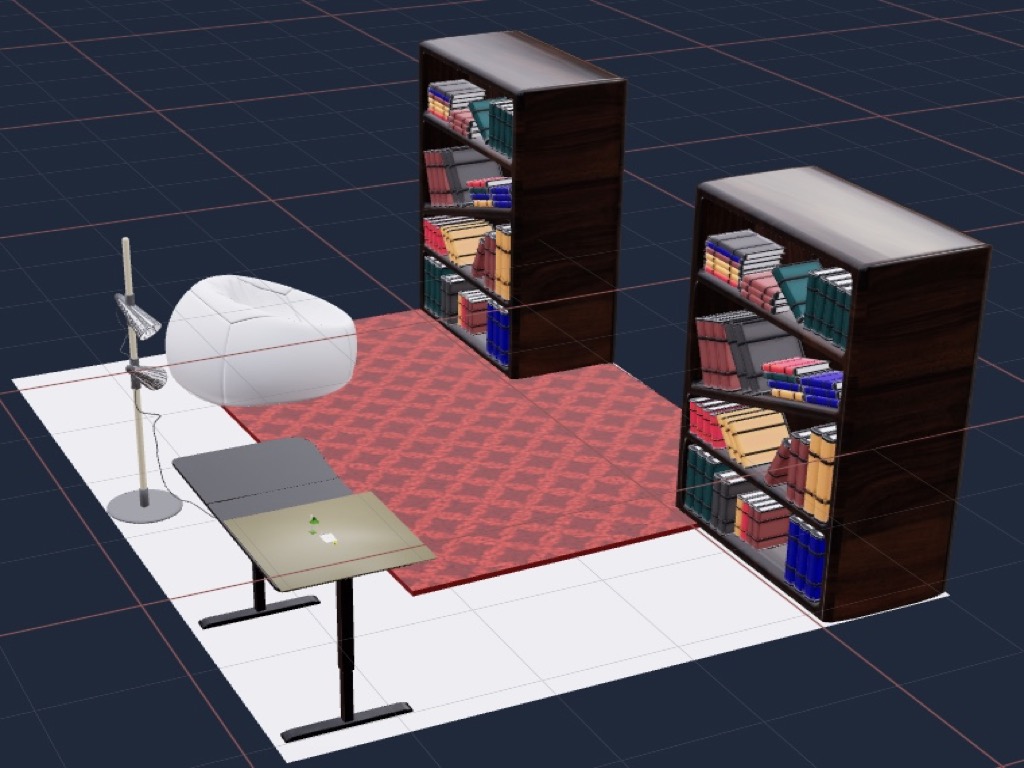}
      \caption{computer room}
      \label{fig:gl4}
  \end{subfigure}
  \caption{Example spaces generated by RoboLayout}
  \label{fig:scene_results}
\end{figure*}

Figure ~\ref{fig:scene_results} presents representative scenes generated by \textit{RoboLayout}. The results demonstrate that the system:
\begin{itemize}
    \item Consistently positions large furniture elements adjacent to walls while satisfying physical feasibility constraints
    \item Preserves semantically appropriate inter-object spacing in accordance with predefined layout policies
    \item Produces coherent object orientations that conform to common indoor design norms
    \item Correctly models hierarchical spatial relationships (e.g., decorative objects placed on supporting furniture) while maintaining structural realism
\end{itemize}

\subsection{Loss Curve}

\begin{figure}[htbp]
\centering
\begin{subfigure}[b]{\columnwidth}
\centering
\includegraphics[width=\textwidth]{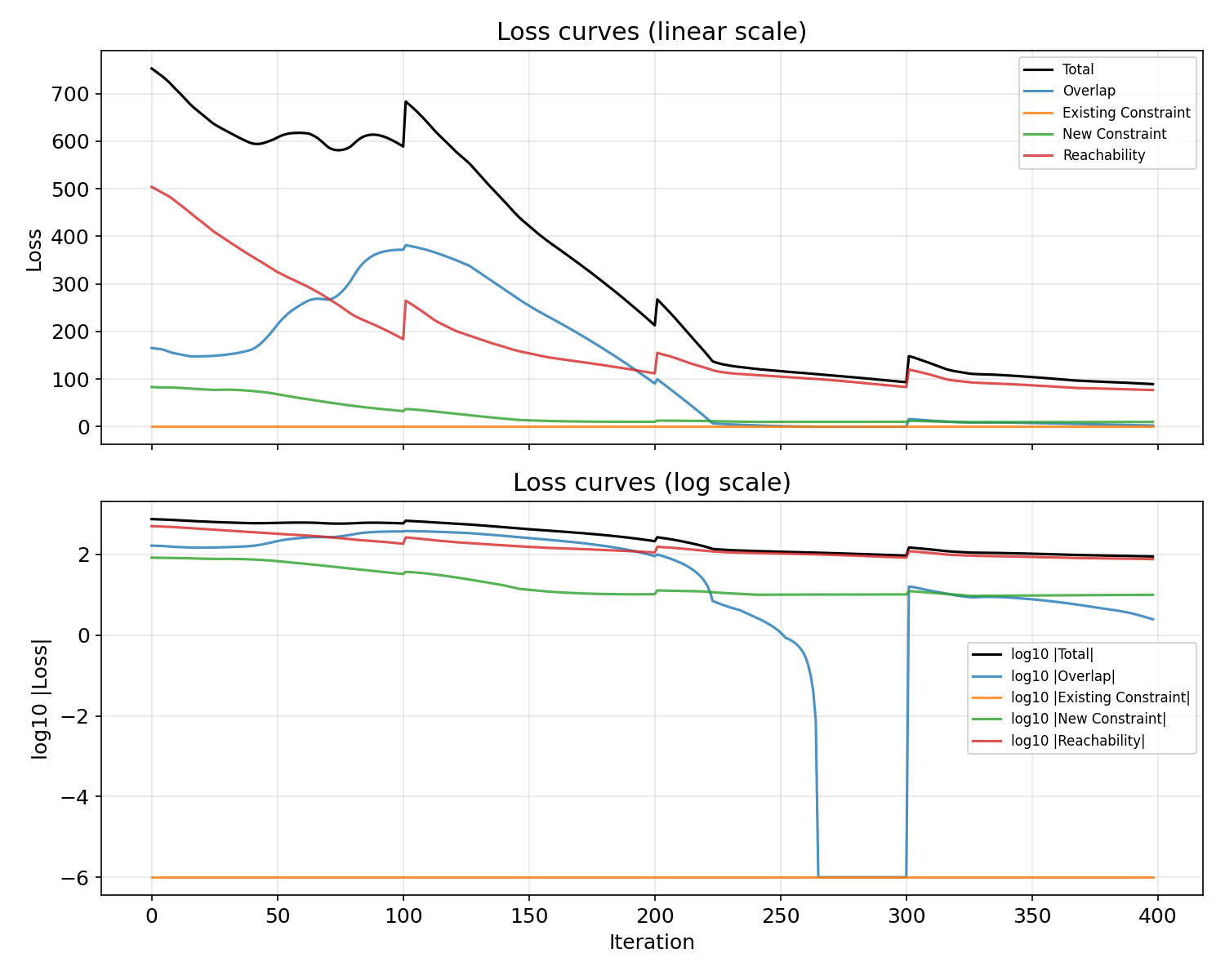}
\end{subfigure}
\caption{Loss Curve}
\label{fig:loss}
\end{figure}

Figure ~\ref{fig:loss} contains two subplots that summarize the \emph{gradient descent optimization} of the layout solver.  Each iteration is one step of gradient descent: the solver computes the loss, backpropagates to get gradients with respect to the layout parameters (3D positions and rotations of objects), then updates those parameters via an optimizer (e.g. Adam).  This is \emph{not} training of a neural network and no model weights are learned from data.  The curves show the loss function minimized by gradient descent for a single scene layout problem. The two panels are:
\begin{itemize}
  \item The \textbf{top} panel shows the evolution of several loss terms in \emph{linear} scale.
  \item The \textbf{bottom} panel shows the same quantities in \emph{logarithmic} scale (base~10) applied to the absolute value of each loss.
\end{itemize}

At each optimization iteration $i \in \{0,1,\dotsc,N-1\}$ the solver stores a dictionary where some keys (e.g. reachability) may be absent if the corresponding term is disabled.  Each entry can be a PyTorch tensor or a Python scalar; for plotting it is always converted to a floating--point number.

\section{ Future Work}

Future research directions include the following:
\begin{itemize}
     \item Integrating robot motion planning directly into the optimization process, replacing the current walking-robot–based clearance constraint. This would enable feasibility-aware layouts that account for kinematic reachability, collision-free trajectories, and task execution constraints, resulting in more realistic and deployable solutions.
     \item Extending the current robot reachability model from a planar, radius-based formulation to a full 3D representation by incorporating vertical reach constraints. This would allow more accurate modeling of robot capabilities and improve the validity of generated layouts in real-world environments.
    \item Investigating alternative optimization strategies beyond gradient-based methods currently used for soft constraint optimization. In particular, hybrid discrete–continuous or unified optimization techniques (e.g., mixed-integer optimization or sampling-based methods) could improve convergence robustness and better handle non-convex, combinatorial layout constraints.
    \item Extending self-consistency beyond the current constraint-level filtering: (1) strengthening conflict resolution and redundancy removal in the constraint set, and (2) adopting self-consistent decoding by sampling several candidate programs and selecting the one that achieves the best solver objective (e.g., total loss or feasibility), thereby improving reliability of generated layouts.

\end{itemize}

\section{Conclusion}

This paper presents RoboLayout, an extension of LayoutVLM that integrates explicit agent-aware reachability into differentiable 3D layout optimization. By treating the “robot” as a general embodied agent, including robots, humans, or animals, RoboLayout generates layouts that are not only semantically coherent but also physically navigable and actionable by the intended agent. In addition, a local refinement stage selectively re-optimizes only problematic object placements while keeping the remainder of the scene fixed, improving convergence efficiency and layout stability without increasing global optimization cost. Experimental results demonstrate that RoboLayout preserves strong semantic alignment, improves optimization robustness, and produces agent-feasible indoor layouts across diverse scene configurations, bringing 3D scene generation closer to practical deployment in real-world environments.

\phantomsection
\section{References}
\label{sec:references}
\label{ref:1}
[1] Fan-Yun Sun, Weiyu Liu, Siyi Gu, Dylan Lim, Goutam Bhat, Federico Tombari
, Manling Li, Nick Haber, Jiajun Wu, Layoutvlm: Differentiable Optimization of 3D Layout via Vision-Language Models, 2025 \\
\label{ref:2}
[2] K. Black, N. Brown, D. Driess, A. Esmail, M. Equi, C. Finn, N. Fusai, L. Groom, K. Hausman, B. Ichter, S. Jakubczak, T. Jones, L. Ke,S. Levine, A. Li-Bell, M. Mothukuri, S. Nair, K. Pertsch, L. X. Shi, J. Tanner, Q. Vuong, A. Walling, H. Wang, and U. Zhilinsky, “$\pi$0: A vision-language-action flow model for general robot control,”arXiv 2410.24164, 2024.\\
\label{ref:3}
[3] M. M. Soliman, E. Ahmed, A. Darwish, and A. E. Hassanien, “Artificial intelligence powered metaverse: analysis, challenges
and future perspectives,” Artificial Intelligence Review, vol. 57,no. 2, p. 36, 2024.\\
\label{ref:4}
[4] D. Liu, J. Zhang, A.-D. Dinh, E. Park, S. Zhang, and C. Xu, “Generative physical AI in vision: A survey,” arXiv 2501.10928, 2025.\\
\label{ref:5}
[5] K. Chang, C. Cheng, J. Luo, S. Murata, M. Nourbakhsh, and Y. Tsuji, “Building-GAN: Graph-conditioned architectural volumetric design generation,” in ICCV, 2021 \\
  \label{ref:6}
[6]W. Zhao, Y. Cao, J. Xu, Y. Dong, and Y. Shan, “DI-PCG: diffusionbased efficient inverse procedural content generation for highquality 3D asset creation,” arXiv 2412.15200, 2024\\
\label{ref:7}
[7] W. Wu, L. Fan, L. Liu, and P. Wonka, “Miqp-based layout design for building interiors,” Computer Graphics Forum, 2018.\\
\label{ref:8}
[8] Y. Li, O. Vinyals, C. Dyer, R. Pascanu, and P. W. Battaglia, “Learning deep generative models of graphs,” arXiv 1803.03324, 2018\\
\label{ref:9}
[9] I. J. Goodfellow, J. Pouget-Abadie, M. Mirza, B. Xu, D. Warde-Farley, S. Ozair, A. C. Courville, and Y. Bengio, Generative adversarial networks, in NIPS, 2014.
\label{ref:10}
[10] J. Ho, A. Jain, and P. Abbeel, “Denoising diffusion probabilistic models,” in NeurIPS, 2020.\\
\label{ref:11}
[11] P. Kan and H. Kaufmann, “Automatic furniture arrangement ´ using greedy cost minimization,” in VR, 2018 \\
\label{ref:12}
[12] Y. Zhao, K. Lin, Z. Jia, Q. Gao, G. Thattai, J. Thomason, and G. S. Sukhatme, “LUMINOUS: indoor scene generation for embodied AI challenges,” arXiv 2111.05527, 2021.\\
\label{ref:13}
[13] M. Fisher, D. Ritchie, M. Savva, T. A. Funkhouser, and P. Hanrahan, “Example-based synthesis of 3D object arrangements,” ACM
TOG, vol. 31, no. 6, pp. 135:1–135:11, 2012.\\
\label{ref:14}
[14] S. Zhang, S. Zhang, W. Xie, C. Luo, Y. Yang, and H. Fu, “Fast 3D indoor scene synthesis by learning spatial relation priors of objects,” IEEE TVCG, vol. 28, no. 9, pp. 3082–3092, 2022.\\
\label{ref:15}
[15] S. Dasgupta, A. Gupta, S. Tuli, and R. Paul, “ActNeRF: Uncertainty-aware active learning of nerf-based object models for robot manipulators using visual and re-orientation actions,” in IROS, 2024.\\
\label{ref:16}
[16] W. Feng, W. Zhu, T. Fu, V. Jampani, A. R. Akula, X. He, S. Basu, X. E. Wang, and W. Y. Wang, “LayoutGPT: Compositional visual planning and generation with large language models,” in NeurIPS, 2023. \\
\label{ref:17}
[17] K. Bhat, N. Khanna, K. Channa, T. Zhou, Y. Zhu, X. Sun, C. Shang, A. Sudarshan, M. Chu, D. Li, K. Deng, J. Fauconnier, T. Verhulsdonck, M. Agrawala, K. Fatahalian, A. Weiss, C. Reiser, R. K. Chirravuri, R. Kandur, A. Pelaez, A. Garg, M. Palleschi, J. Wang, S. Litz, L. Liu, A. Li, D. Harmon, D. Liu, L. Feng, D. Goupil, L. Kuczynski, J. Yoon, N. Marri, P. Zhuang, Y. Zhang, B. Yin, H. Jiang, M. van Workum, T. Lane, B. Erickson, S. Pathare, K. Price, A. Singh, and D. Baszucki, “Cube: A roblox view of 3D intelligence,” arXiv 2503.15475, 2025.\\
\label{ref:18}
[18] Y. Yang, J. Lu, Z. Zhao, Z. Luo, J. J. Q. Yu, V. Sanchez, and F. Zheng, “LLplace: The 3D indoor scene layout generation and
editing via large language model,” arXiv 2406.03866, 2024. \\
\label{ref:19}
[19] R. Fu, Z. Wen, Z. Liu, and S. Sridhar, “AnyHome: Openvocabulary generation of structured and textured 3D homes,”
in ECCV, 2024. \\
\label{ref:20}
[20] Yandan Yang1, Baoxiong Jia1, Shujie Zhang1,  Siyuan Huang1, All-in-One 3D Scene Synthesis with an Extensible and Self-Reflective Agent
 1State Key Laboratory of General Artificial Intelligence, BIGAI 2Tsinghua University \\
 \label{ref:21}
  [21] Y. Yang, F. Sun, L. Weihs, E. VanderBilt, A. Herrasti, W. Han, J. Wu, N. Haber, R. Krishna, L. Liu, C. Callison-Burch, M. Yatskar, A. Kembhavi, and C. Clark, “Holodeck: Language guided generation of 3D embodied AI environments,” in CVPR, 2024. \\
 \label{ref:22}
 [22] M. Zhou, J. Hou, C. Luo, Y. Wang, Z. Zhang, and J. Peng, “SceneX: procedural controllable large-scale scene generation via largelanguage models,” in AAAI, 2025.\\
 \label{ref:23}
 [23]  R. Aguina-Kang, M. Gumin, D. H. Han, S. Morris, S. J. Yoo, A. Ganeshan, R. K. Jones, Q. A. Wei, K. Fu, and D. Ritchie, “Openuniverse indoor scene generation using LLM program synthesis and uncurated object databases,” arXiv 2403.09675, 2024 \\
 \label{ref:24}
 [24] Ian W. Eisenberg, The Unified Control Framework: Establishing a Common Foundation for Enterprise AI Governance, Risk Management and Regulatory Compliance , 2025 \\
 \label{ref:25}
 [25] Emmanouil Papagiannidis a , Patrick Mikalef, Kieran Conboy b Responsible artificial intelligence governance: A review and research framework Author links open overlay panel, The Journal of Strategic Information Systems, June 2025, 101885 \\
\label{ref:26}
[26] Automated architectural space layout planning using a physics-inspired generative design framework, Z. Li, S. Li, G. Hinchcliffe, N. Maitless, N. Birbilis, 2024.\\
\label{ref:27}
[27] Retail Innovation Through AI: Benefits and Liabilities , W. Michael Schuster, Simon E. Corrigan, 2025.
 
\end{document}